\documentclass[10pt, a4paper, twocolumn]{article}

\usepackage[lmargin=0.6in, rmargin=0.6in, tmargin=0.6in, bmargin=1.0in]{geometry}
\usepackage{fancyhdr}
\usepackage{changepage}
\usepackage{titlesec}
\usepackage{indentfirst}
\usepackage{hyperref}
\usepackage{blindtext}
\usepackage{romannum}
\usepackage{enumitem}
\usepackage{float}
\usepackage{multirow}
\usepackage{bm}
\usepackage{graphicx}
\usepackage{caption}
\usepackage{subcaption}
\usepackage{placeins}
\usepackage{newtxtext}
\usepackage{newtxmath}

\graphicspath{{./images/}}

\setenumerate{noitemsep}

\titlespacing{\section}{0pt}{10pt}{0pt}
\titlespacing{\subsection}{0pt}{0pt}{0pt}

\hypersetup{
    colorlinks=true,
    urlcolor=blue
}

\title{\textbf{Optical Flow Based Motion Detection for Autonomous Driving}}
\author{Ka Man Lo \\ 
    \small kamanphoebe@gmail.com \\\\ 
    \small \url{https://github.com/kamanphoebe/MotionDetection.git}
}
\date{}

\begin{document}

\pagenumbering{arabic}
\maketitle

\begin{center}
    \textbf{Abstract}
\end{center}
\vspace{-.25cm}
\begin{adjustwidth}{20pt}{20pt}
    \textit{
    \indent Motion detection is a fundamental but challenging task for autonomous driving. In particular scenes like highway, remote objects have to be paid extra attention for better controlling decision. Aiming at distant vehicles, we train a neural network model to classify the motion status using optical flow field information as the input. The experiments result in high accuracy, showing that our idea is viable and promising. The trained model also achieves an acceptable performance for nearby vehicles. Our work is implemented in PyTorch. Open tools including nuScenes, FastFlowNet and RAFT are used. Visualization videos are available at \url{https://www.youtube.com/playlist?list=PLVVrWgq4OrlBnRebmkGZO1iDHEksMHKGk} .
    }
\end{adjustwidth}

\section{\large Introduction}
    
    Motion detection, or moving object detection, is a computer vision related technique for detecting the physical movement of an object relative to its background. It is widely used in various areas like smart homes, surveillance and security, and also plays a crucial role in autonomous driving. To make better future plan on controlling during driving, vehicles need to monitor the road condition well. Careful inspection for faraway environment is required for scenes that allow high-speed driving like highways or quiet roads. However, the perception range of lidar and radar sensors are not always far enough to cover distant objects and thus computer vision based methods should be applied under these circumstances. Traditional methods of motion detection rely on the difference of pixels between frames. Therefore, detecting motion in the distance, especially those in radial direction, is a challenging issue since they are usually just a few pixels changes.
    
    Optical flow estimation is a commonly used technique in motion detection tasks for providing velocity information. It is calculated based on the brightness constancy constraint, supposing the timestamps of two consecutive frames are close enough that the brightness of the same positions in real world will remain unchanged. In this paper, we use different algorithms to obtain optical flow field information of vehicles in between 30 to 70 meters from the nuScenes \cite{nuscenes2019} dataset, and feed them into neural network ResNet18 \cite{DBLP:journals/corr/HeZRS15} as inputs. The model then outputs the binary prediction of motion status, i.e., still or moving. Our experiments show that the moving targets are successfully detected with a high correct rate. We also use the trained model to infer nearby vehicles and obtain a reasonable accuracy.
    
    The rest of the paper is organized as follows: Section \Romannum{2} gives a brief review of relevant works. Section \Romannum{3} demonstrates the framework of our work, followed by the experimental details and results in Section \Romannum{4}. Finally, conclusions and possible future work are presented in Section \Romannum{5}.

\section{\large Related Work}

    In this section, basic approaches regarding motion detection are firstly reviewed, together with specific topics relevant to our work, namely small object motion detection and autonomous driving. After that, we dive deeper into optical flow algorithms.

\subsection{\normalsize Motion detection}

    Traditional methods used in detecting motion can be mainly divided into four categories: background subtraction, frame difference, temporal difference and optical flow estimation. A previous review done by Manchanda and Sharma \cite{7508161} includes works using these approaches to detect motion for general purposes. The works they list were published from 2009 to 2015. There is also some afterwards improvement based on the basic methods in recent years, such as  \cite{8864015}\cite{8328591}\cite{9429516}\cite{8833206}. 
    
    For recent works about motion detection in autonomous driving, both  \cite{9443120} and \cite{8076832} make use of CRF related model. The former targets at a specific range while the latter jointly feeds disparity map and optical flow field as model input. Our work is inspired by \cite{8569744}. Yet we concentrate solely on motion classification for bounding boxes so far while their work combines object detection and motion segmentation. Besides, we exploit the original optical flow information instead of converting it to RGB images so as to prevent normalization in this process and preserve numerical precision. More details about our implementation are stated in Section \Romannum{3}.
    
    Focusing on the topic of small object motion detection, existing works usually focus on insects, like \cite{DBLP:journals/corr/abs-1805-00342}\cite{rs13040653}, which entirely differ from autonomous driving in appearance and background.

\subsection{\normalsize Optical flow}

\begin{figure*}[t!]
        \centering
        \begin{subfigure}{.8\textwidth}
            \centering
            \includegraphics[width=\textwidth]{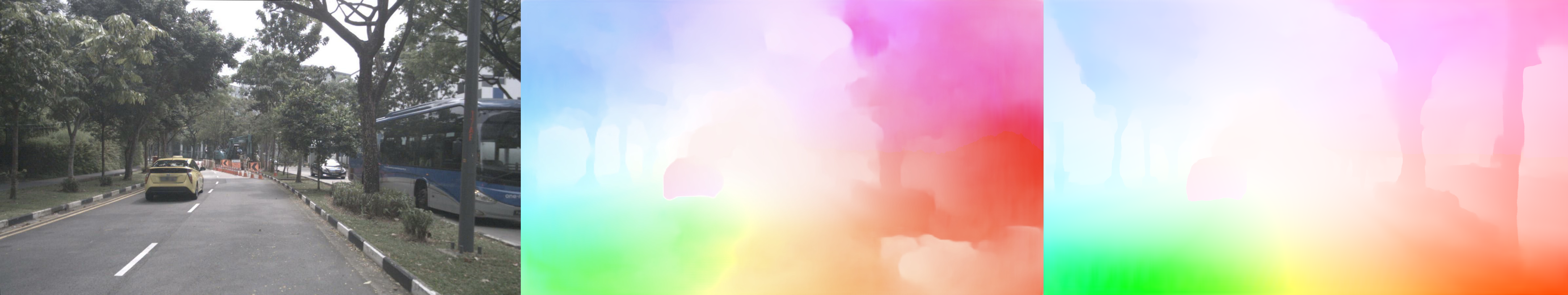}
        \end{subfigure}
        \par\smallskip
        \begin{subfigure}{.8\textwidth}
        \centering
        \includegraphics[width=\textwidth]{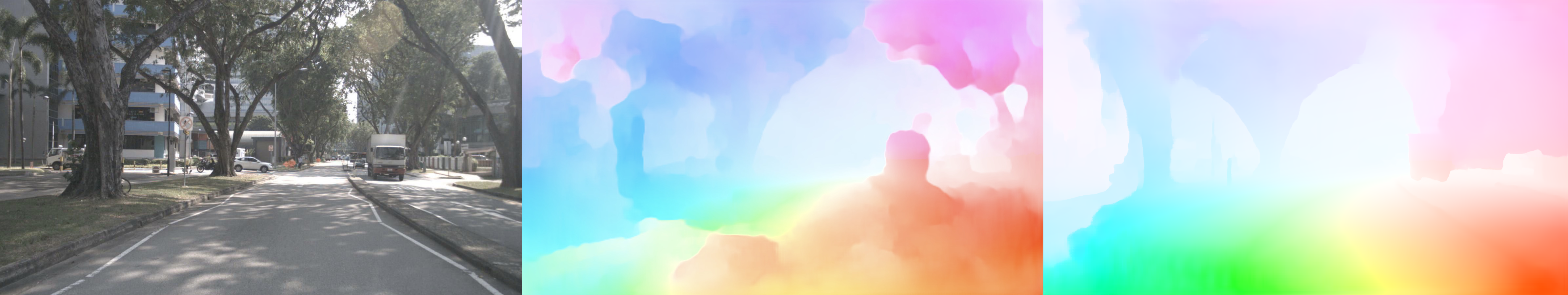}
        \end{subfigure}
    \caption{Examples of optical flow predictions on the nuScenes dataset. From left to right: the preceding raw image of a keyframe pairs, flow visualization of FastFlowNet and RAFT.}
    \label{fig:optical_flow}
    \end{figure*}

    FastFlowNet \cite{Kong:2021:FastFlowNet} and RAFT \cite{DBLP:journals/corr/abs-2003-12039} achieve state-of-the-art speed and accuracy respectively for estimating optical flow field. FastFlowNet is $10\times$ faster than RAFT while RAFT obtains a F1 error of $5.10\%$ on the KITTI \cite{Geiger2013IJRR} dataset, which is only half of the value of FastFlowNet. The two algorithms are used and are compared with each other in our work. Example inferences of FastFlowNet and RAFT using the same raw image pair are depicted in Figure \ref{fig:optical_flow}.

\section{\large Method}

    The framework of our work is presented in this section, starting with the pipeline and followed by details of labeling and data preprocessing.

\subsection{\normalsize Pipeline}

    The overview of our work is outlined below:
    
    \begin{enumerate}[noitemsep, topsep=0pt]
        \item Select keyframe pairs that contain target vehicles from nuScenes
        \item Generate optical flow field for all keyframe pairs via FastFlowNet or RAFT
        \item Label the objects as still or moving by estimating their velocity  
        \item Extract optical flow information of objects within the modified 2D bounding boxes after some preprocessing and feed them into the neural network
        \item Train a binary classifier from scratch using ResNet18 along with some necessary adjustments of layers
    \end{enumerate}
    
\subsection{\normalsize Labeling}

    Data of 2D bounding box and binary motion ground truth are recorded in every label. The former is marked by the coordinates \texttt{xmin}, \texttt{xmax}, \texttt{ymin} and \texttt{ymax}, which are simply deduced from the eight corners of the original 3D bounding box by picking the minimum and the maximum of \texttt{x} and \texttt{y}. 
    
    The motion ground truth is decided based on the velocity calculated as
    \begin{equation}
        velocity = \frac{position_{2} - position_{1}}{timestamp_{2} - timestamp_{1}}
    \end{equation}
    where $position$ is given with respect to the global coordinate system. If the absolute value of velocity $\mid velocity \mid \, \geq 2 \, \textrm{m/s} $, then the object is marked as moving or else it is still.  

\subsection{\normalsize Data preprocessing}

    To determine whether an object is moving, we need the optical flow information not only of the object itself, but also of the background around. Therefore, some preprocessing has to be done on the 2D bounding box before inputting to the network, as mentioned in the 4th step of the pipeline. First, the box is reshaped to be a square with side length $\bm{}{max}(width, height)$. Then triple the length of sides and if needed, pad the box with edge values. Finally, cut off data outside the box and resize it to $224\times224$ using bilinear interpolation.

\section{\large Experiments}

    In this section, we first detail the dataset used and experimental setup. Then the results are discussed. The construction and evaluation of a generalized dataset are presented at last.

\subsection{\normalsize Dataset}

    Our model is trained and evaluated on the filtered nuScenes \cite{nuscenes2019} dataset. nuScenes comprises 1000 diverse scenes, covering different locations, time and weather conditions. For simplicity, we exclude scenes of "night", "rain" and "lightning", thereby 604 scenes are remained. We then collect keyframe pairs that contain any of the seven types of vehicles within the specific distance and visibility range as shown in Table \ref{tab:dataset_setting}. After that, optical flow field of the frame pairs is calculated through FastFlowNet or RAFT, and is saved as \texttt{.npy} file. As a result, we obtain 18460 objects in total, while 16467 of them used as training set and 1993 for evaluation. Considering the amount of data is rather small, random horizontal flip with probability 0.5 is performed for data augmentation.
    
    \begin{table}[t!]
    \centering
    \caption{Settings for filtering the nuScenes dataset.}
    \label{tab:dataset_setting}
    \begin{tabular}{c c}
        \hline
         Setting & Description \\
        \hline\hline    
        \multirow{7}{7.5em}{\centering Target categories} & vehicle.car \\ & vehicle.emergency.ambulance \\ & vehicle.emergency.police \\
        & vehicle.truck \\ & vehicle.bus.bendy \\ & vehicle.bus.rigid \\ & vehicle.construction \\
        \hline
        Distance & 30m - 70m \\
        \hline
        Visibility & 80\% - 100\% \\
        \hline
        Sensor & CAM\_FRONT \\
        \hline
    \end{tabular}
    \end{table}
   
\subsection{\normalsize Experimental setup}

    The model architecture we used is chosen to be ResNet18 \cite{DBLP:journals/corr/HeZRS15}. However, since the input is not in the form of RGB images, we have to train the model from scratch instead of applying a pretrained model. Therefore, the number of output channels of the first convolutional layer is modified to be 64 to make the network adapt to our input. The size of final output is also changed to 1, which should be a number within [0, 1]. If the output value greater than 0.5, the object will be classified as moving, otherwise it will be still. The rest of the model structure stay unchanged. Table \ref{tab:hyperparameters} lists the settings of hyperparameters.
    
    \begin{table}[t!]
    \centering
    \caption{Hyperparameter setting for ResNet18 model training.}
    \label{tab:hyperparameters}
    \begin{tabular}{c c c}
        \hline
        Function & Hyperparameter & Setting \\
        \hline\hline    
        Dataloader & Batch size & 128 \\
        \hline
        \multirow{4}{5em}{\centering Optimizer} & Algorithm & SGD \\ & Learning rate & 0.01 \\ & Weight decay & 0.01 \\ & Momentum & 0.9 \\
        \hline
        \multirow{3}{5em}{\centering Scheduler} & Schedule & StepLR \\ & Step size & 10 \\ & Gamma & 0.5 \\
        \hline
    \end{tabular}
    \end{table}

    \begin{figure*}[t!]
        \centering
        \subcaptionbox{Correct predictions}{
            \includegraphics[width=.45\textwidth]{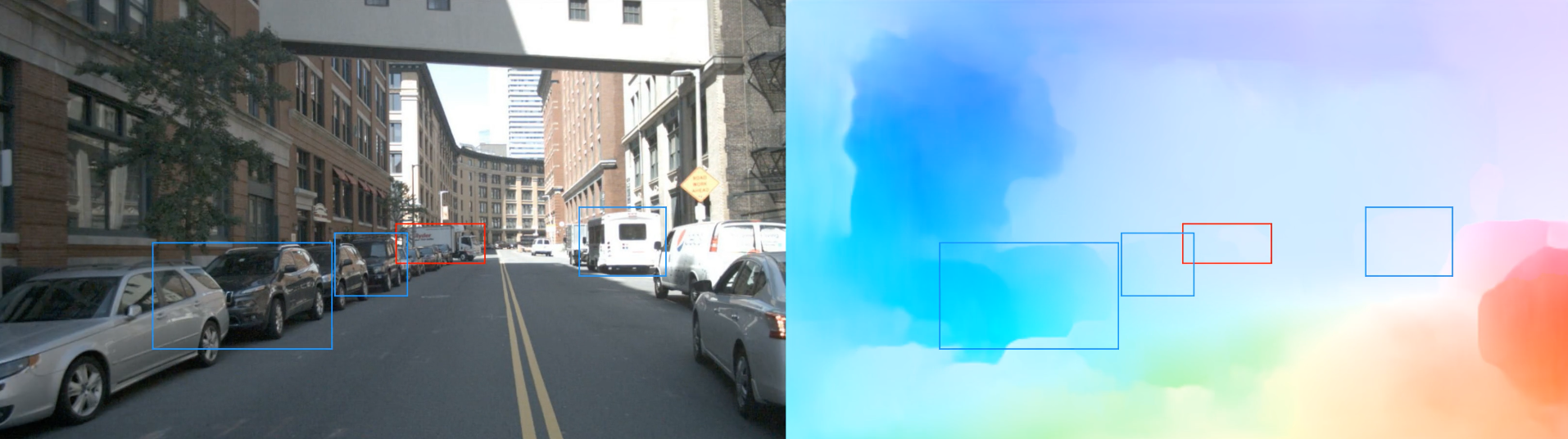}
            \includegraphics[width=.45\textwidth]{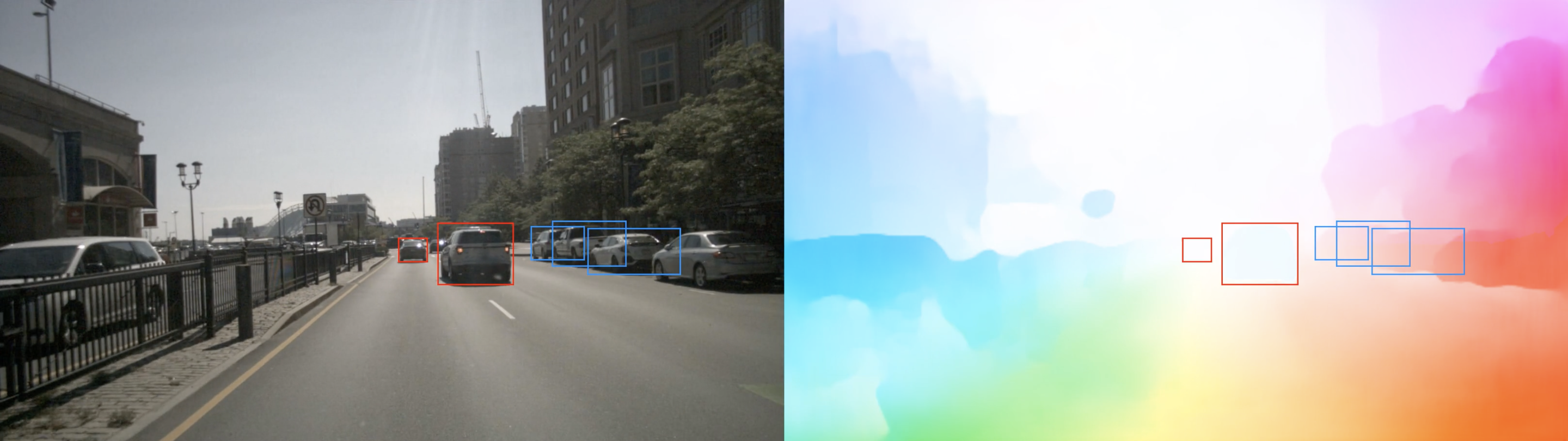}
        }
        \par\smallskip
        \subcaptionbox{Incorrect predictions due to unclear flow}{
            \includegraphics[width=.45\textwidth]{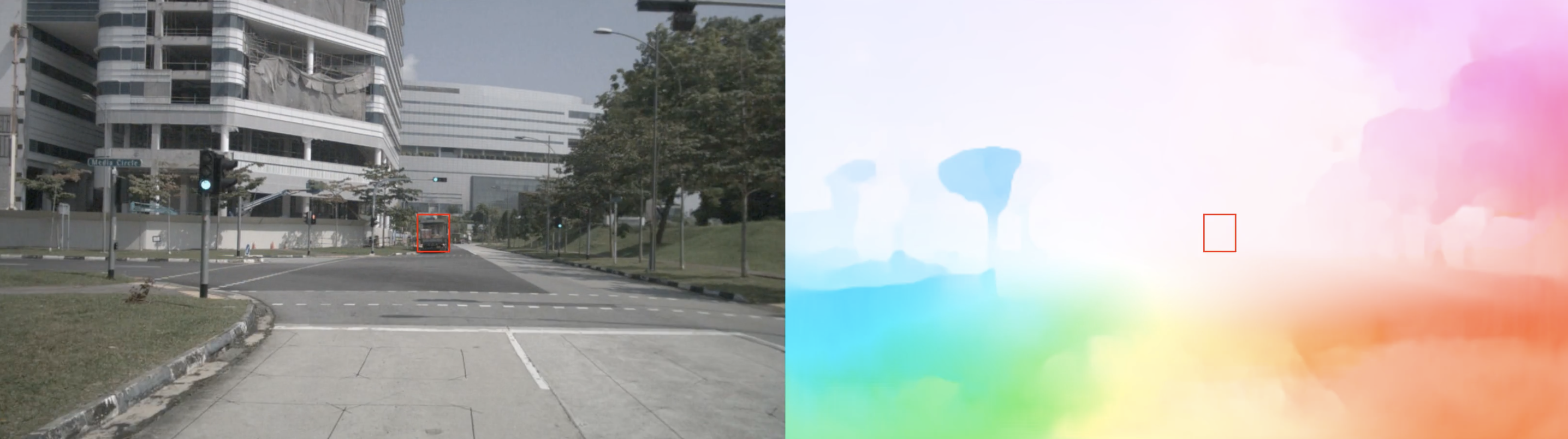}
            \includegraphics[width=.45\textwidth]{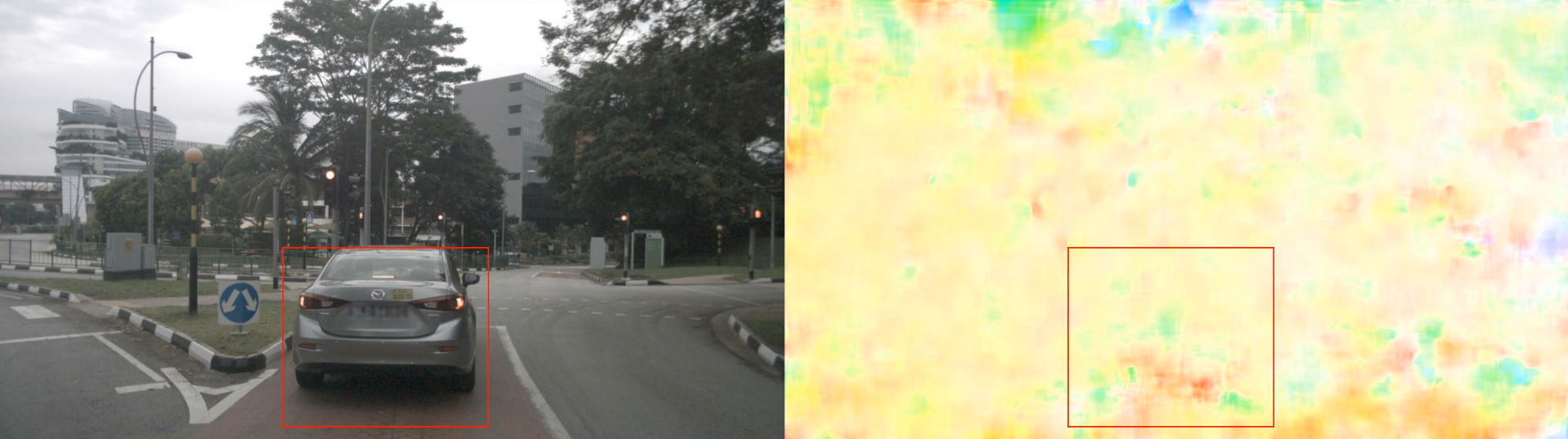}
        }
        \par\smallskip
        \subcaptionbox{Incorrect predictions due to misleading surround}{
            \includegraphics[width=.45\textwidth]{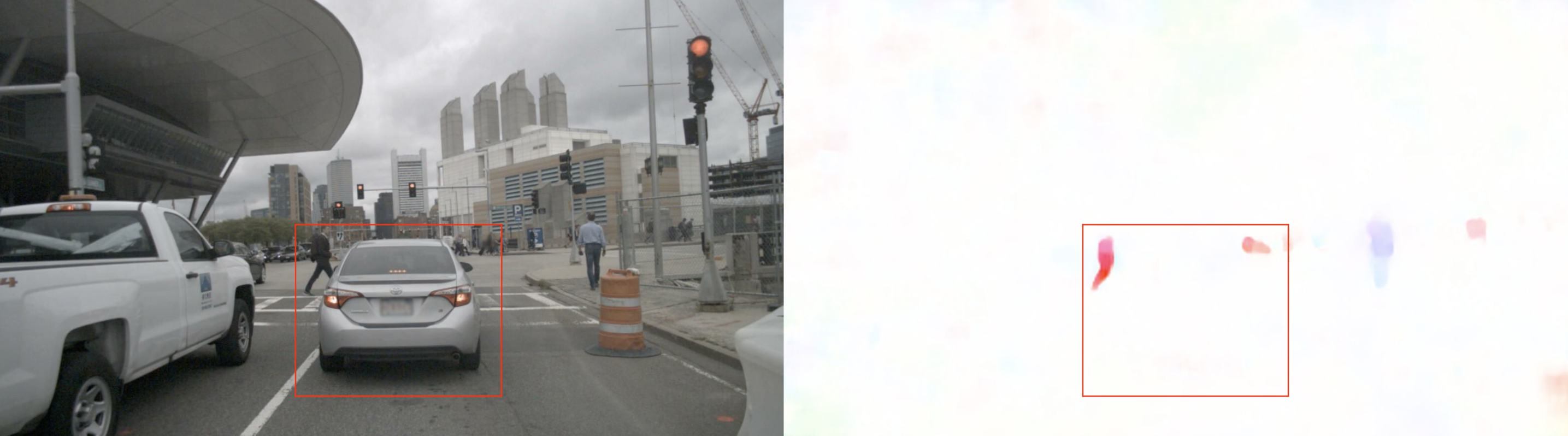}
            \includegraphics[width=.45\textwidth]{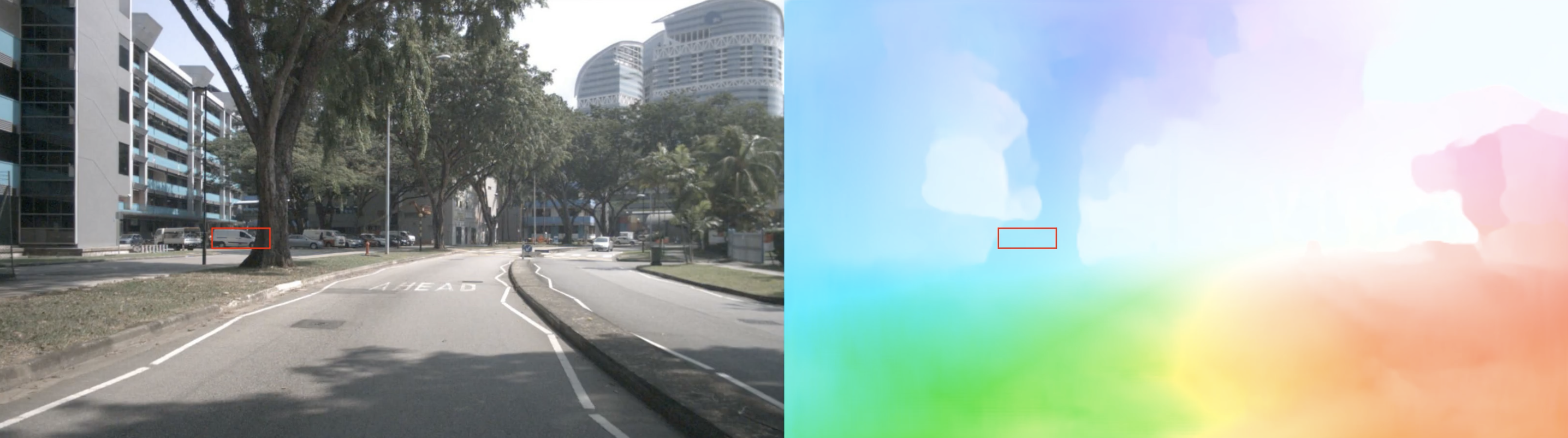}
        }
        \caption{Visualization of inference using the model trained with FastFlowNet input. Blue boxes and red boxes represent still and moving objects respectively. These are the frames of the generalized dataset so nearby objects are also detected.}
        \label{fig:visualization}
    \end{figure*}

\subsection{\normalsize Experimental result}

    The accuracy of our model is greater than 90\%, as shown in Table \ref{tab:result}. The model trained with FastFlowNet input unexpectedly has better performance than the one with input generated by RAFT, which achieves state-of-the-art accuracy for optical flow estimation. Note that it is not fair to compare our results directly with other motion detection methods, since we simplify the data a lot. Nonetheless, the pretty values still reflect the feasibility of our idea of work.
    
    \begin{table}[t!]
    \caption{Quantitative performance of models with optical flow input generated by various algorithms. The pretrained models of both optical flow algorithms are trained on the KITTI \cite{Geiger2013IJRR} dataset.}
    \label{tab:result}
    \centering
    \begin{tabular}{c c c c}
        \hline
        Optical flow & \multirow{2}{3.5em}{\centering F1 (\%)} & \multirow{2}{6em}{\centering Precision (\%)} & \multirow{2}{5em}{\centering Recall (\%)} \\ algorithm & \\
        \hline\hline    
        FastFlowNet & 92.9 & 94.3 & 91.7 \\
        \hline
        RAFT & 89.5 & 89.7 & 89.9 \\
        \hline
    \end{tabular}
    \end{table}
    
    Predictions are visualized for intuitive understanding of the model performance. Several correct and incorrect inferences are depicted in Figure \ref{fig:visualization}. We sum up two main reasons for the wrong classifications:
    
    \begin{itemize}[noitemsep, topsep=0pt]
        \item \textbf{Unclear optical flow for remote or slow objects}. These kinds of objects are always tough to deal with because of the tiny difference in distance in the visual world, hence the unclear flow which confuses the network.
        \item \textbf{Being affected by the background or foreground}. As mentioned in Section \Romannum{3}, the surrounding flow information is also included as part of the input. Therefore, it could be misleading when there are some other objects involved. 
    \end{itemize}

\subsection{\normalsize Generalization}

    We extend our filtered dataset for inference by adding non-keyframes and nearby objects from nuScenes. To start with, optical flow is calculated for frame pairs at 4-frame intervals. Since there is no annotation for non-keyframes in nuScenes, bounding box positions and ground truths of objects are estimated based on the information of their corresponding keyframes. The former are calculated by linear interpolation while the latter remain the same as the closest previous keyframes with respect to the non-keyframes. Vehicles closer than 30 meters, but with visibility requirements as before, are also taken into account. Evaluating on this generalized dataset, the F-score of our model significantly drops to 60\%, as shown in Table \ref{tab:generalization}. Our visualization videos contain inference of generalization.
    
    \begin{table}[h!]
    \caption{Quantitative performance of evaluation on the generalized dataset. The pretrained model of optical flow algorithm is trained on KITTI.}
    \label{tab:generalization}
    \centering
    \begin{tabular}{c c c c}
        \hline
        Optical flow & \multirow{2}{3.5em}{\centering F1 (\%)} & \multirow{2}{6em}{\centering Precision (\%)} & \multirow{2}{5em}{\centering Recall (\%)} \\ algorithm & \\
        \hline\hline    
        FastFlowNet & 60.4 & 63.0 & 62.8 \\
        \hline
    \end{tabular}
    \end{table}

\section{\large Conclusion and future work}

    In this paper, we have investigated the effect of binary motion classification for annotated remote vehicles by inputting optical flow information into a neural network. The experimental result reports that our model can successfully detect the motion and the high accuracy illustrates the great potential of our idea. Cases failed to be correctly inferred are mainly caused by unclear optical flow and misleading surrounding flow information. Our trained model is not so applicable to nearby objects yet the performance might be dramatically enhanced if the model is trained with them. There is still much room for improvement for our work:
    
    \begin{itemize}[noitemsep, topsep=0pt]
    \item Remove the strict rules for filtering data to adapt to concrete use
    \item Train the optical flow model from scratch (by self-supervised learning), rather than apply pretrained models
    \item Construct an end-to-end classification network architecture, leaving the middle stages regarding generating optical flow field to be implicit
    \end{itemize}

\section*{\large Acknowledgement}
    
    The code for generating optical flow field information is based on the corresponding original projects, FastFlowNet and RAFT.

\bibliographystyle{plain}
\bibliography{refs}

\end{document}